%% file: main.tex
\documentclass[10pt,twocolumn,letterpaper]{article}
\pdfoutput=1
\usepackage{cvpr}
\usepackage{times}
\usepackage{epsfig}
\usepackage{graphicx}
\usepackage{amsmath}
\usepackage{amssymb}
\usepackage{multirow}
\usepackage{algorithm}
\usepackage[noend]{algpseudocode}
\usepackage{balance}
\usepackage{tabularx}
\usepackage{array}

\makeatletter
\def\BState{\State\hskip-\ALG@thistlm}
\DeclareMathOperator*{\argmaxA}{arg\,max}

\makeatother

\cvprfinalcopy 

\def\cvprPaperID{ }

\ifcvprfinal\fi
\begin{document}

	\title{Correlation Propagation Networks for Scene Text Detection} 
	
	\author{Zichuan Liu$^1$, Guosheng Lin$^1$, Wang Ling Goh$^1$, Fayao Liu$^2$, Chunhua Shen$^2$ and Xiaokang Yang$^3$ \\
			$^{1}$Nanyang Technological University, Singapore\\
			$^{2}$University of Adelaide, Australia\\
			$^{3}$Shanghai Jiao Tong University, China}
	
	\maketitle
	\thispagestyle{empty}

	\input{abstract2.tex}
	\input{introduction2.tex}
	\input{related_works2.tex}

	\input{method3.tex}
	\input{experiments.tex}
	\input{conclusion.tex}
	\input{acks.tex}
	
	\newpage
	\clearpage
	\balance
	
	{\small
		\bibliographystyle{ieee}
		\bibliography{egbib}
	}
	
	\newpage
	\clearpage
	
	\input{appendix.tex}
\end{document}

%% file: abstract2.tex
\begin{abstract}
	In this work, we propose a novel hybrid method for scene text detection namely Correlation Propagation Network (CPN). It is an end-to-end trainable framework engined by advanced Convolutional Neural Networks. Our CPN predicts text objects according to both top-down observations and the bottom-up cues. Multiple candidate boxes are assembled by a spatial communication mechanism call Correlation Propagation (CP). The extracted spatial features by CNN are regarded as node features in a latticed graph and Correlation Propagation algorithm runs distributively on each node to update the hypothesis of corresponding object centers. The CP process can flexibly handle scale-varying and rotated text objects without using predefined bounding box templates. Benefit from its distributive nature, CPN is computationally efficient and enjoys a high level of parallelism. Moreover, we introduce deformable convolution to the backbone network to enhance the adaptability to long texts. The evaluation on public benchmarks shows that the proposed method achieves state-of-art performance, and it significantly outperforms the existing methods for handling multi-scale and multi-oriented text objects with much lower computation cost.
\end{abstract}

%% file: introduction2.tex
\section{Introduction}
The scene text detection has drawn much attention in both academic communities and industry due to its ubiquitous applications in the real world. It is an essential component in modern information retrieval and advanced automobile system. \par

The mainstream scene text detection methods are the variants of the general object detection methods based on Convolutional Neural Networks (CNNs) \cite{lecun1998gradient}. They treat words or text lines as individual objects which can be detected by the CNN with learned patterns. Benefit from the powerful representation capability of the CNNs and large amounts of training data, these methods show robustness in a stable environment where text objects have regular shapes and aspect ratios. However, in the case dominated by the scale- and aspect-ratio- varying texts, these methods could fail due to the limited field of view of CNNs and improper reference box design. This problem can be mitigated by enlarging receptive of the CNNs or introducing multi-scale features to encode more informative contexts \cite{lin2017feature,liu2016ssd}. But, as will be demonstrated in the Sect. \ref{sect:challenge}, they still suffer from degradation in detecting extreme long and oriented text objects. \par

In this paper, we analyze the challenges of bounding box regression in existing text detection methods and propose a novel framework namely Correlation Propagation Network (CPN) which is capable of detecting multi-oriented and scale-varying text objects. It is inspired by the recently proposed Markov Clustering Network (MCN) \cite{liu2018mcn} which considers the text detection in a bottom-up manner. In comparison, our method is a hybrid method which explores both global information and local correlation of an object. Our CPN predicts bounding boxes independently at each anchor location and merges the predicted box via distributive Correlation Propagation among adjacent anchors. The Correlation Propagation (CP) mechanism allows the anchors to progressively obtain the perception of possible object centers by propagating local correlation spatially via weighted connections between anchors, and the final bounding boxes are output by assembling boxes with the same object center. The CP mechanism is built on top of existing neural networks and allows end-to-end optimization. Instead of computing the gradients of the iterative CP process, a recursive back-propagation algorithm is developed to approximate the gradient which remarkably speedups the training while maintains the performance. Moreover, by analyzing the pattern of local connections, we design a Greedy Path Selection (GPS) algorithm to further accelerate the inference with lower complexity and memory footprint. Finally, we enhance the capability of encoding long text objects by introducing adaptive sampling provided by deformable convolution \cite{dai2017deformable}. \par

The CPN is naturally flexible without using predefined reference templates and is extremely suitable for the scenarios where texts appear in arbitrary size and orientation. Both the training and inference are efficient and highly parallelizable. We evaluate our method on public benchmarks and prove its effectiveness and robustness to the large variety of scales, aspect ratios, and orientations. The contribution of this work is summarized as follows:
\begin{itemize}
	\item A flexible Correlation Propagation (CP) mechanism for scene text detection is proposed which adaptively merges bounding boxes for scale-varying and multi-oriented text;
	\item A recursive algorithm is proposed to compute the gradients of CP operation. It is computationally efficient and can be fully paralleled on GPUs;
	\item A highly paralleled Greedy Path Selection (GPS) algorithm is proposed to accelerate the CP process in the testing phase without loss of accuracy;
	\item The deformable convolution \cite{dai2017deformable} is further introduced to text detection and performance improvement is empirically demonstrated; 
	\item Our method shows superior performance in detecting multi-oriented scene text object with the precision of 89.8, recall of 82.7, F-score of 86.1 on MSRA-TD500 \cite{zhang2016multi} and frame rate of 10 FPS with very deep ResNet-50 backbone \cite{he2016deep}.
\end{itemize}

%% file: related_works2.tex
\section{Related Works}

Retrieving texts in natural scenes has been widely investigated for years, and great progresses have been achieved in both academic community and industry in terms of robustness and flexibility. According to different application scenarios, the reported scene text detection methods can be summarized into four categories. \par

\textbf{Component-based} These are the methods that retrieve word or text lines by recognizing individual characters and their relations in between \cite{zhu2016scene,zamberletti2014text,shi2013scene,neumann2012real}. The most well-known method \cite{huang2013text} applies Stroke Width Transform (SWT) or Maximally Stable Extremely Region (MSER) as a character extractor to generate character candidates, which will be refined and connected into words or text lines. This kind of methods are effective but not well-scalable in scenarios with complex background due to the limitation of handcrafted features. Additionally, these methods are usually inefficient due to their low parallelism. \par

\textbf{Detection-based} In detection-based methods, text retrieval are formulated as an object detection problem, which can be implemented in character level and word level. A detection window is applied to equal-stride image areas to classify whether there exists any text and regress the corresponding bounding boxes. In the recent years, detection-based methods engined by Convolutional Neural Networks (CNNs) are the most successful \cite{neumann2016real,wang2012end,huang2013text,huang2014robust,yao2012detecting,wang2010word,zhang2016multi,zhang2015symmetry,bissacco2013photoocr,jaderberg2016reading,gupta2016synthetic}. They have become the mainstream solutions for the task due to its end-to-end trainable characteristics. However, the performance may degrade remarkably in the case where the statistics of text geometry is highly varying. \par  

\textbf{Segmentation-based} These methods inherits from the semantic segmentation, which generate dense heatmap to reveal the occurrence of texts \cite{girshick2016region,dai2017fused,dai2017fused}. Since the high-density classification is adopted, the segmentation-based methods can be aware of the text geometry and produce elaborated boundary. However, the performance highly depends on the post-processing to remove false-positives and aggregate candidates. In addition, the multistage processing flow requires exhaustive tuning and results in long processing time. \par

\textbf{Spatial Connectionist} The reported methods for scene text detection is still facing the problem in dealing with scale-varying and multi-oriented texts. Different from general objects, texts have unconstrained length, shape and orientation. A text instance should be considered as a combination of text components rather than an individual object. However, the correlation between components may not be well-captured by either the detection-based or segmentation-based methods. Work \cite{tian2016detecting} realizes text detection as a sequential prediction which is supported by the proposed spatial recurrent neural network. But it is limited in detecting horizontal text lines and the parallelism is restricted by the sequential structure of RNNs. Works \cite{liu2018mcn,shi2017detecting} introduce link prediction to classic object detection frameworks to capture the correlation between text components. Work \cite{liu2018mcn} enjoys better scalability since it does not require predefined box templates and have faster inference speed. \par

Our method can be categorized as a spatial connectionist method and is inspired by the recent proposed Markov Clustering Network (MCN) \cite{liu2018mcn}. Both are common in adopting the graphical model to identify text instance but different in basic idea, inference algorithm, and learning. First, CPN is a hybrid method that produces text candidates base one top-down template matching and assembles them according to predicted object centers. While MCN is a bottom-up method utilizes local correlation to cluster highly correlated regions. Second, MCN is a centralized method while CPN is a distributive message passing mechanism. Third, CPN does not consist of any heuristic parameter in the learning algorithm and have much lower training complexity.

%% file: method3.tex
\section{Method}

\subsection{Challenges in Scene Text Detection}
\label{sect:challenge}
\begin{figure}[]
	\centering
	\includegraphics[width=\linewidth]{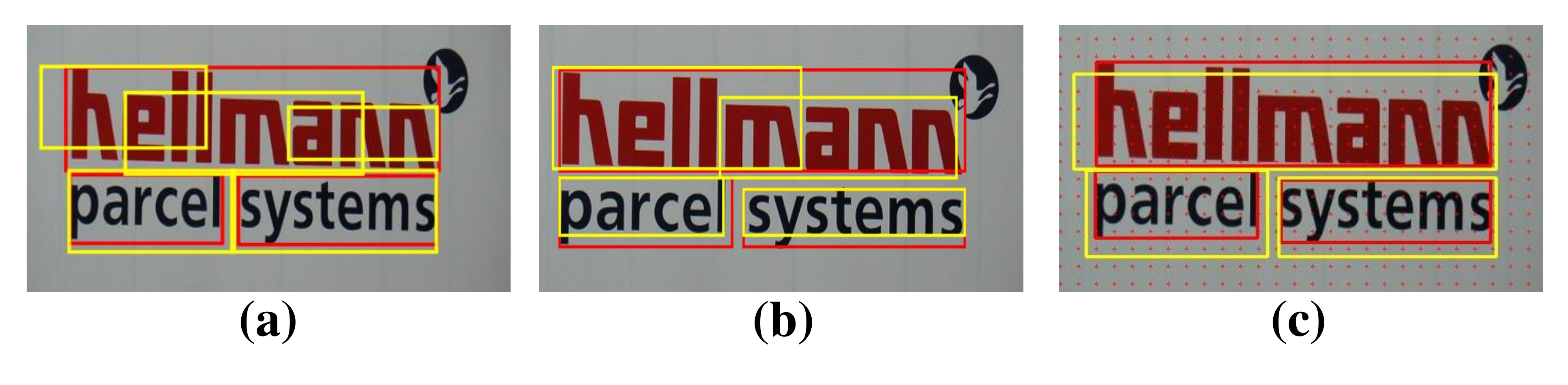}
	\caption{(a) Bounding boxes proposed by Faster R-CNN; (b) Bounding boxes proposed by Faster R-CNN with deformable convolution; (c) Bounding boxes proposed by CPN. Rectangles in red are ground-truth boxes and those in yellow are predictions.}
	\label{fig:bad_example}
\end{figure}

Before introducing our method, we discuss the problem occurred when using general object detection methods to detect texts. The general object detection flow often suffers from remarkable performance degradation \cite{tian2016detecting} when directly applied to scene text detection. As shown in Fig. \ref{fig:bad_example} (a), a Faster R-CNN model trained on scene text datasets cannot perform well when detecting long text lines. Although applying deformable convolution \cite{dai2017deformable} is able to improve the performance, the model still fails to provide a correct bounding box proposal as shown in Fig. \ref{fig:bad_example} (b). It is because the CNN model with a limited size of receptive field may not be able to perceive the whole scene text object due to its large aspect ratio and varying orientation. CNN fails to encode sufficient information to capture the long distant dependency. Thus, a flexible prediction mechanism is desired to ensemble information produced from multiple observations. A prediction of the object center is made according to both current observation and the neighboring predictions. This spatially conditional prediction is realized by Correlation Propagation Network (CPN). In the next subsections, we will introduce our CPN method with detail discussion of the Correlation Propagation (CP) mechanism. The data labeling method and training algorithm will be provided in the following subsections.\par

\subsection{Correlation Propagation Networks}

\begin{figure}[]
	\centering
	\includegraphics[width=\linewidth]{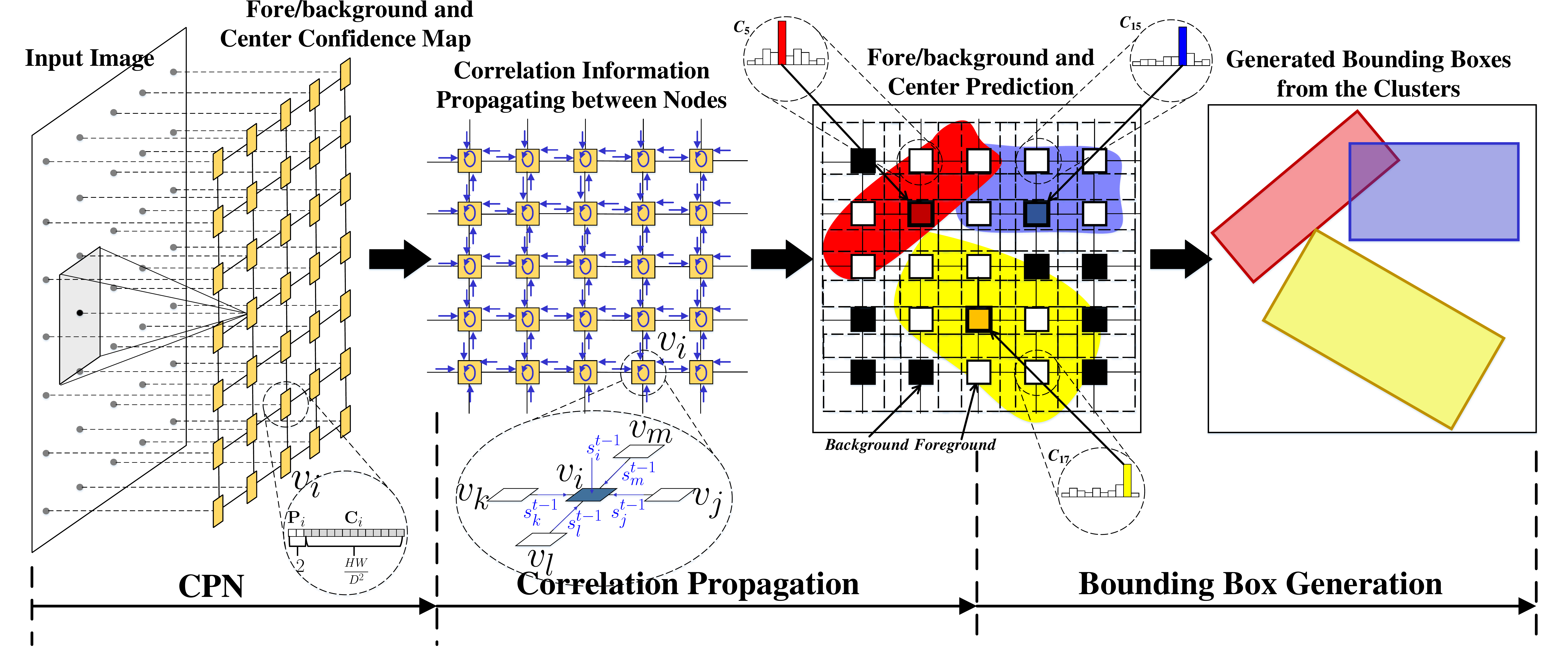}
	\caption{Architecture of Correlation Propagation Networks.}
	\label{fig:arch}
\end{figure}

As shown in Fig. \ref{fig:arch}, a Correlation Propagation Network (CPN) interprets an image with size of $H\times W$ to a latticed graph $G(V,E)$ with $\frac{H}{D}\times \frac{W}{D}$ nodes \footnote{$D$ denotes the down-sampling factor.}. Every node is associated with a feature vector extracted from the corresponding subregion in the original image. For each node $v_i$, CPN predicts fore/background confidence $\mathbf{P}_i\in \mathbb{R}^{2}$ and object center confidence $\mathbf{C}_i \in \mathbb{R}^{\frac{HW}{D^2}}$ from the extracted feature. The predicted $\mathbf{P}_i$ and $\mathbf{C}_i$ will be further constructed to a spatial fore/background confidence map $\mathbf{P} \in \mathbb{R}^{\frac{H}{D}\times \frac{W}{D} \times 2}$ and center confidence map $\mathbf{C} \in \mathbb{R}^{\frac{H}{D}\times \frac{W}{D} \times \frac{HW}{D^2}}$. \par

In CPN, the fore/background prediction acts as a mask on the center confidence map $\mathbf{C}$ to remove the background nodes. The center confidence map $\mathbf{C}$ is generated by a differentiable Correlation Propagation (CP) operation. The CP mechanism allows nodes to communicate with each other to produce hypothesis of possible object centers. One can obtain instance-level object information by grouping nodes sharing the same center prediction. Since nodes are associated with a subregion in the original image, the object geometry can be recovered from the grouped nodes. By applying Principle Component Analysis (PCA) to the grouped nodes, one can provide robust bounding box generation which is insensitive to variation of the size and orientation. \par

Correlation Propagation (CP) mechanism is an iterative process where every node autonomously collects local information from its four nearest neighbors to obtain a global view of an image. As shown in Fig. \ref{fig:arch}, $v_i$ keeps updating its confidence $\mathbf{C}_i$ by iteratively receiving information from its neighbors $\mathcal{N}(i) := \{j,k,l,m\}$. The update rule is illustrated by 
\begin{align} \label{eq:cp}
\mathbf{C}_i^t = \sum_{\mathcal{N}(i)} s_{\mathcal{N}(i)} \cdot \mathbf{C}_{\mathcal{N}(i)}^{t-1} + s_i\cdot \mathbf{C}_i^{t-1},
\end{align}
where $\mathbf{C}_i^t$ denotes the center confidence of node $v_i$ at $t$ step and $s_{(\cdot)}$ is a local correlation measurement between $v_i$ and its neighbors $v_{N(i)}$. Nodes have larger $s$ are highly correlated and they can share more information with each other during the correlation propagation, which helps to identify the boundary between objects. Thus, $s_i$ and $s_{\mathcal{N}(i)}$ can be also interpreted as the weights that determine how to combine the confidence vectors of current and adjacent nodes at previous step.  \par

\begin{figure}[] 
	\centering
	\includegraphics[width=\linewidth]{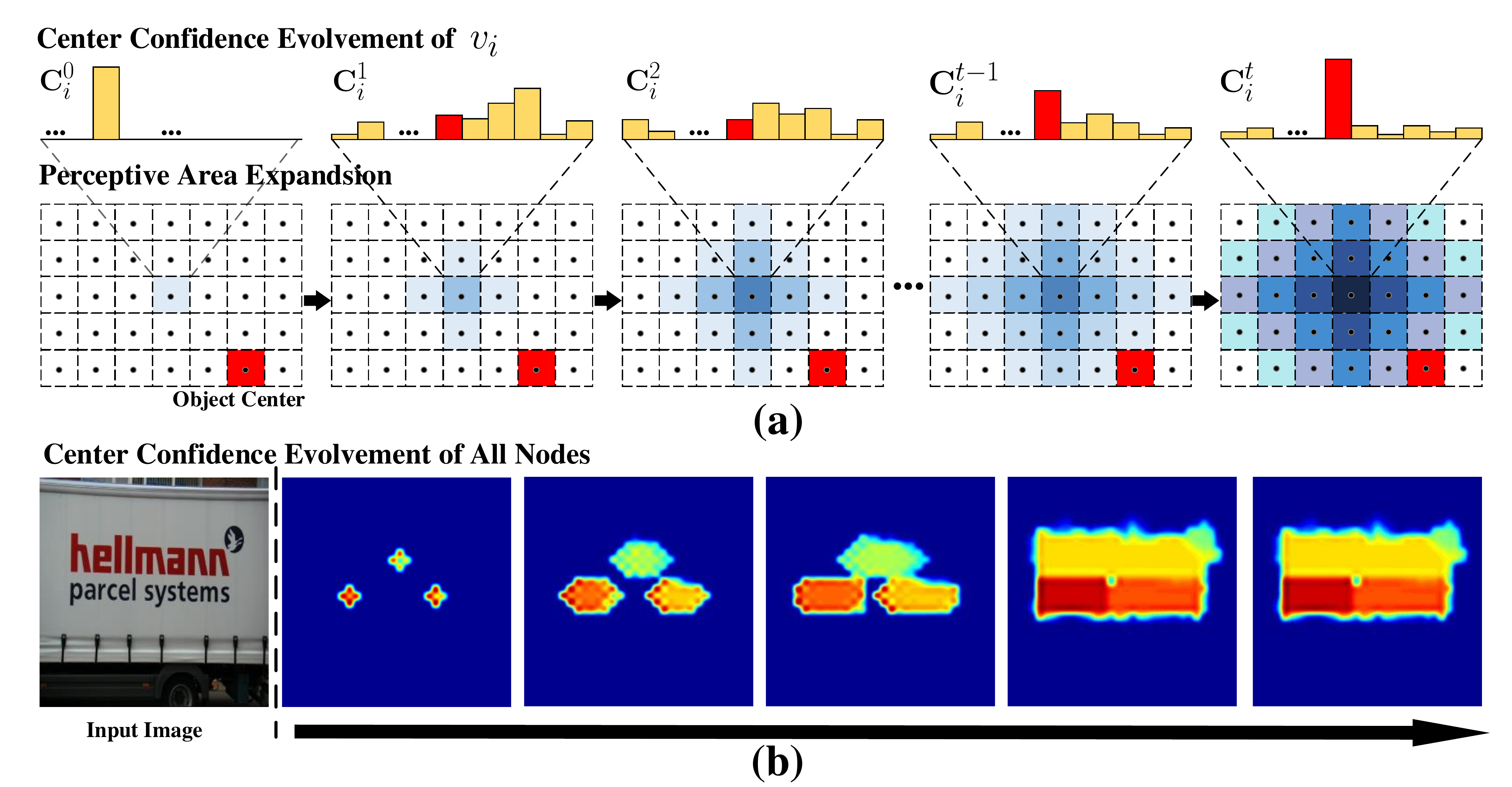}
	\caption{Correlation Propagation: (a) Visualization of center confidence evolvement and perceptive area expansion of $v_i$; (b) Object center confidence evolvement of all nodes.}
	\label{fig:cor_prop}
\end{figure}

As the CP process independently runs at each node, the local correlation will be propagated throughout the whole graph. For a specific node $v_i$, it conceptually expands its perceptive area during the correlation propagation by collecting messages from its neighbors and renew its hypothesis about according object center. The first row of Fig. \ref{fig:cor_prop} (a) visualizes the evolvement of the center confidence $\mathbf{C}_i$. At the beginning, $\mathbf{C}_i$ is initialized as a one-hot vector with $i$-the entry labeled as 1, and becomes smooth and uniform at the next several iterations. As the iteration executed, $\mathbf{C}_i$ becomes more and more peaked and converges to a vector with one or more dominated values whose indices indicate the possible object center nodes for $v_i$. The perceptive area expansion of $v_i$ is illustrated in the second row of Fig. \ref{fig:cor_prop} (a).  Once the perceptive area reaches the potential object center, the existence of the center will be reflected on $\mathbf{C}_i$ with high response at the corresponding entry. \par

The CP process is iteratively boosting the elements in $\mathbf{C}_i$ corresponding to center nodes and suppressing the others. Nodes closer to the corresponding centers will converge earlier, and the confidence of object centers appears to be growing spatially from the potential center nodes. The dynamics of confidence growing is demonstrated in Fig. \ref{fig:cor_prop} (b) with a real example. For illustration, the confidence map of the center nodes are overlapped and resized to the original size of the input image. As shown in the last five columns of Fig. \ref{fig:cor_prop} (b), increasing number of nodes notice the existence of corresponding centers. The information of centers gradually spreads to the associated nodes belonging to the same object. Combined with the fore/background mask, the instance-level outline of a text object can be identified.\par 

\subsection{Bounding Box Assembling}
For object detection, bounding box regression is considered as a general approach to indicate the location of an object. However, it may not be perfect in scene text detection due to the existence of long text object which may not covered by the reference template. The Correlation Propagation is used to assemble the imperfect bounding box prediction into perfect one. Here, we adopt the bounding box prediction approach proposed in work \cite{zhou2017east} to predict orientation and the distance to top, bottom, left, right side of the bounding box at each anchor location. The predicted boxes are merged if they share the same center prediction. Final detection results are produced by applying NMS to the merged boxes.

\subsection{Learning Correlation Propagation}
In this section, we introduce the learning algorithm for CPN to perform correlation propagation between nodes.
\subsubsection{Data Labeling}
In CPN, two types of labels are used, the fore/background mask and the object center label. A fore/background mask $\mathcal{M}_o \in \{0,1\}^{\frac{M}{D}\times \frac{N}{D}}$ is generated from the ground-truth bounding box according to its geometry coverage. The object center label $\mathcal{M}_c \in \mathbb{R}^{\frac{MN}{D^2}\times \frac{MN}{D^2}}$ describes the ground-truth object center of each node. We first find the center of a bounding box and assign the node closest to the object center as the center node. $\mathcal{M}_c $ is constructed by labeling all nodes with the index of the corresponding center node. It is notable that a node can have multiple centers since it could be covered by multiple bounding boxes. For a node $v_i$ covered by $n$ bounding boxes, the corresponding entrance of $\mathcal{M}_c(i)$ will be labeled as $\frac{1}{n}$. 

\subsubsection{Loss Function}
Given a fore/background mask $\mathcal{M}_o$, object center label $\mathcal{M}_c$ and predicted bounding box regression, we formulate a multi-task learning problem to train CPN, which is illustrated as follows:
\begin{align}
\begin{split} &\mathcal{C} = \frac{D^2}{MN} \sum_{i} [\mathcal{L}_o(i)  + \alpha\cdot\mathcal{L}_c(i) + \beta\cdot\mathcal{M}_o(i)\cdot\mathcal{L}_r(i)] \end{split}, \\
\begin{split} &\mathcal{L}_o(i) = -\mathcal{M}_o(i)\cdot\ln (\mathbf{P}_i) \end{split}, \\
\begin{split} &\mathcal{L}_c(i) = -\mathcal{M}_c(i)\cdot\ln (\mathbf{C}_i)\end{split}, \\
\begin{split} &\mathcal{L}_r(i) = L_1(\mathbf{R}_i)\end{split}.
\end{align}
The cost function $\mathcal{C}$ consists of fore/background loss $\mathcal{L}_o$, center prediction loss $\mathcal{L}_c$ weighted by a parameter $\alpha$ and box regression loss weighted by parameter $\beta$ and ground-true foreground mask $\mathcal{M}_o$. $\mathcal{L}_o$ and $\mathcal{L}_c$ are cross-entropy loss decided by predicted $\mathbf{P}_i, \mathbf{C}_i$ and corresponding ground-true labels $\mathcal{M}_o$ and $\mathcal{M}_c$. $L_1(\cdot)$ represents the soft L1 loss proposed in \cite{girshick2015fast}.

\subsubsection{Learning Algorithm for Correlation Propagation}
In this subsection, we derive the learning algorithm for Correlation Propagation. As illustrated in Eq. \ref{eq:cp}, the hypothesis $\mathbf{C}_i$ at node $v_i$ is the weighted sum of hypothesis of current and neighborhood nodes at the last time-step. Eq. \ref{eq:cp} can be recursively expanded as 
\begin{align}
\begin{split} 
\mathbf{C}_i^t &= \sum_{\mathcal{N}(i)} s_{\mathcal{N}(i)} \cdot \mathbf{C}_{\mathcal{N}(i)}^{t-1} + s_i\cdot \mathbf{C}_i^{t-1} \\
&= \sum_{\mathcal{N}(i)} s_{\mathcal{N}(i)} \cdot [\sum_{\mathcal{N}(\mathcal{N}(i))} [\cdots[\sum_{\mathcal{N}\cdots(i)} s_{\mathcal{N}\cdots(i)} \cdot \mathbf{C}_{\mathcal{N}\cdots(i)}^{0}  \\ 
& \ \ \ \ \ \ \ \ \ \ \ \ \ \ + s_{\mathcal{N}\cdots(i)}\cdot \mathbf{C}_{\mathcal{N}\cdots(i)}^{0}]\cdots] \\
& \ \ \ \ \ \ \ \ \ \ \ \ \ \ + s_{\mathcal{N}(\mathcal{N}(i))}\cdot \mathbf{C}_{\mathcal{N}(\mathcal{N}(i))}^{t-2}] + s_i\cdot \mathbf{C}_i^{t-1}
\end{split},
\end{align}
which is fully decided by local correlation prediction $s_V$ \footnote{$s_V$ denotes local correlation measurements of nodes in set $V$.}. Thus, the scene text detection problem is interpreted as predicting the local correlation measurement $s_V$. To derive the corresponding gradients, we rewrite the cross-entropy loss of center prediction of a node $v_i$ as:
\begin{align}
\label{eq:loss}
\begin{split}
\mathcal{L}_c(i) & = -\mathcal{M}_c(i)\cdot\ln (\mathbf{C}_i^t) \\
& = -\mathcal{M}_c(i) \cdot\ln (\sum_{\mathcal{N}(i)} s_{\mathcal{N}(i)} \cdot \mathbf{C}_{\mathcal{N}(i)}^{t-1} + s_i\cdot \mathbf{C}_i^{t-1}) \\
& = \sum_{j\in\{j:\mathcal{M}_c(i,j) \neq 0\}} -\mathcal{M}_c(i,j) \\
& \cdot\ln [\sum_{\mathcal{N}(i)} s_{\mathcal{N}(i)} \cdot \mathbf{C}_{\mathcal{N}(i)}^{t-1}(j) + s_i\cdot \mathbf{C}_i^{t-1}(j)]
\end{split}, 
\end{align}
and the gradients with respect to $s_i$ and $s_{\mathcal{N}(i)}$ are represented by
\begin{align}
\begin{split} \label{eq:grad_1} & g_{s_i} = \frac {\partial \mathcal{L}_c}{\partial s_i} = -\sum_{j} \frac{\mathcal{M}_c(i,j) \cdot \mathbf{C}^{t-1}_i(j)}{\mathbf{C}^{t}_i(j)}\end{split}, \\
\begin{split} \label{eq:grad_2}& g_{s_{\mathcal{N}(i)}} = \frac {\partial \mathcal{L}_c}{\partial s_{\mathcal{N}(i)}} = -\sum_{j} \frac{\mathcal{M}_c(i,j)\cdot \mathbf{C}^{t-1}_{\mathcal{N}(i)}(j)}{\mathbf{C}^{t}_i(j)}\end{split}.
\end{align}
As shown in Eq. \ref{eq:grad_1} and Eq. \ref{eq:grad_2}, the gradients with respect to $s_{V}$ is decided by the confidence values of ground-true center node at $t$-th step $\mathbf{C}^{t}_i(j)$ and at $(t-1)$-th step $\mathbf{C}^{t}_{(\cdot)}(j)$. Thus, computing the gradients with respect to $s_{V}$ is transformed into calculating the confidence value of the ground-true center node for each $v_i$. Since $\mathbf{C}^{t}_i(j)$ recursively depends on its adjacent nodes, one approach is running Eq. \ref{eq:cp} at every node for $T$ iterations before computing $g_{s_i}$ and $g_{s_{\mathcal{N}(i)}}$. However, this approach has a complexity of $\mathcal{O}(\frac{TMN}{D^2})$ and the selection of $T$ is empirical. This iterative process remains to be task-dependent and will greatly slow down the training. Recalling the CP process illustrated in Fig. \ref{fig:cor_prop} (b), the information of an object center spreads from the object center to its adjacent nodes and then to distant nodes. Nodes closer to the object center will actually recognize their centers earlier and the corresponding $\mathbf{C}_i(j)$ will converge in advance. Therefore, the gradients with respect to $s_{V}$ of closer to center nodes can be computed and backpropagated during the CP process.  Based on above analysis, a recursive algorithm (Algorithm \ref{alg:train_opt_1}) is developed to compute $g_{s_i}$ and $g_{s_{\mathcal{N}(i)}}$ efficiently. \par

\begin{figure}[b]
	\centering
	\includegraphics[width=\linewidth]{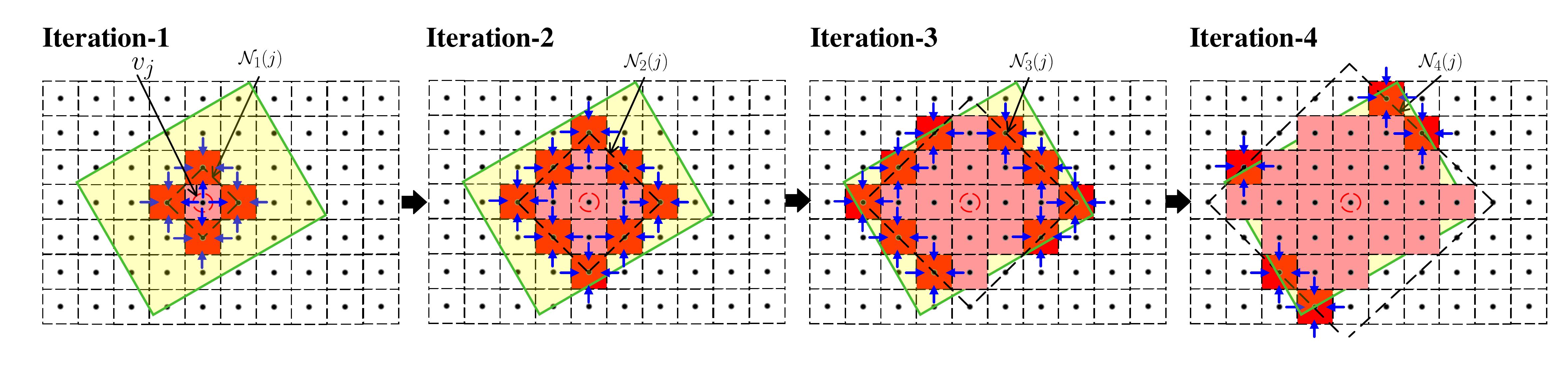}
	\caption{Visualization of Recursive Gradient Computation: Nodes labeled in red are activated in the current iteration. Nodes labeled in pink were activated in previous iterations.}
	\label{fig:cp_grad}
\end{figure}

For the sake of clearness, we first introduce the definition of k-step neighbors of a node. For a center node $v_j$, its $k$-step neighbors denoted as $V_{\mathcal{N}_{k}(j)}$ are the nodes from which one can move (horizontally or vertically) to $v_j$ by $k$ steps. Fig. \ref{fig:cp_grad} shows 1- to 4-step neighbors of a center node $v_j$. Additionally, $V_{c} \subseteq V$ denotes all the ground-true object centers and $V_{\mathcal{N}}:=\{V_{\mathcal{N}_{k}(j)}: j\in V_{c}, k \in K_j\}$ represents the $k$-step neighbors for each center. Given a center node $v_j \in V_c$ and nodes $V_o(j)$ covered by corresponding bounding box, the algorithm updates the confidence value $\mathbf{C}_{\mathcal{N}_{k}(j)}(j)$, $\mathcal{L}_c(\mathcal{N}_{k}(j))$, $g_{s_{\mathcal{N}_{k}(j)}}$ and $g_{s_{\mathcal{N}({\mathcal{N}_{k}(j)})}}$ of the nodes which are the intersection between $V_{\mathcal{N}_{k}(j)}$ and $V_o(j)$ by running Eq. \ref{eq:cp}, \ref{eq:loss}, \ref{eq:grad_1}, \ref{eq:grad_2} iteratively. The recursive algorithm eliminates the empirical $T$ selection and only has a complexity of $\mathcal{O}(\sum_{j\in V_c} \sum_{k=1}^{K_j}|\mathcal{N}_k(j)|)$, making CPN to be more general in scenarios with various text geometries. The gradients computation for one bounding box is visualized in Fig. \ref{fig:cp_grad}. Notably, above process can run successively for each $v_j \in V_c$ to support nodes shared by different bounding boxes. The whole recursive algorithm is summarized in Algorithm \ref{alg:train_opt_1}.

\input{cp_grad}

\subsection{Accelerating Inference with Greedy Path Selection}
\begin{figure}[t]
	\centering
	\includegraphics[width=\linewidth]{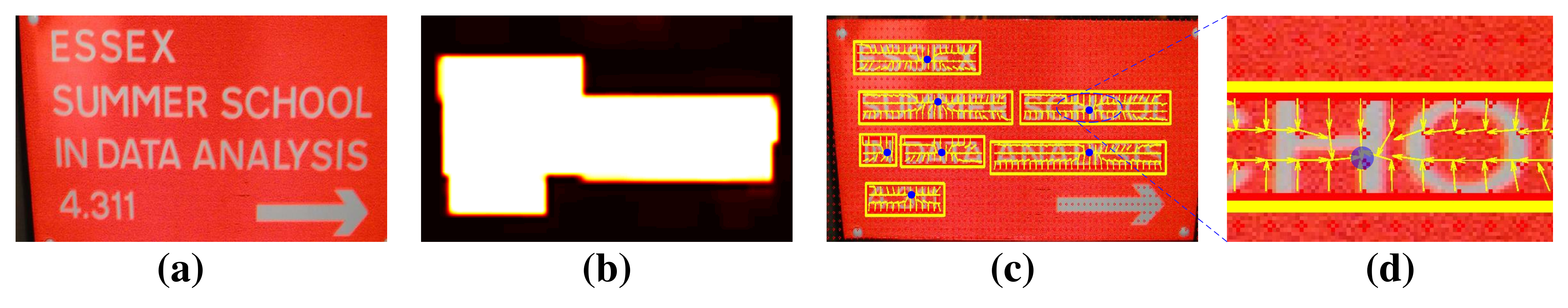}
	\caption{(a) Input image; (b) Predicted foreground heatmap; (c) Visualization of combined vector $\omega$ of each node; (d) The partial enlarge view of $\omega$ around center node.}
	\label{fig:s_example}
\end{figure}

In this subsection, we propose a Greedy Path Selection (GPS) algorithm which further accelerates the CP process. It comes from the observation that the predicted $s_{\mathcal{N}_i}$ are consistently pointing to the interior of an object. More precisely, if we treat $s_{\mathcal{N}_i}$ as the magnitudes of four vectors pointing out to four directions (top, down, left and right) and visualize the combined vector $\omega$, we find that they consistently point inward as demonstrated in Fig. \ref{fig:s_example}. Moreover, starting at an arbitrary node within an object region, one can reach its corresponding center node with high probability by following the path constructed by $\omega$. \par

Based on the analysis above, the GPS algorithm is designed to make nodes $v$ independently search their object center greedily. Each node continuously seeks its next-hop node by selecting a neighborhood node with largest $s_{\mathcal{N}_i}$ until it is trapped to a specific node $Tr_i$. $Tr_i$ is labeled as a candidate of object center for $v_i$. In most cases, nodes within the same object will be trapped at the same $Tr_i$. But the ambiguity still exists in the case where multiple candidates are close to each other. For this case, we apply original CP process only in the area covering these center candidates to check if they can be merged to one center. In practice, this two-stage approach works well to identify object centers and is robust to close small objects and inaccuracy of $s$ prediction. In terms of computational complexity, the original CP process have the complexity of $\mathcal{O}(\frac{TMN}{D^2})$, while the GPS enjoys the complexity of $\mathcal{O}(\sum_{j\in V_c} \sum_{k=1}^{K_j} k\cdot|\mathcal{N}_k(j)|)$. Since the CP process is used in a small sub-graph, it does not dominate the computation and the total complexity roughly remains $\mathcal{O}(\sum_{j\in V_c} \sum_{k=1}^{K_j} k\cdot|\mathcal{N}_k(j)|)$. Rather depending on the size of the input image, the computational complexity of GPS-accelerated inference algorithm is decided by the number of objects contained in an image and their sizes, which is much faster than the original CP-based inference. In our experiment, we show that the GPS-accelerated inference is 24$\times$ faster than the pure CP-based inference algorithm. The GPS accelerated inference algorithm is summarized in Algorithm \ref{alg:gps}.

\vspace{0pt}
\input{gps}

%% file: cp_grad.tex
\begin{algorithm}
	\caption{Recursive Gradient Computation}
	\label{alg:train_opt_1}
	\begin{algorithmic}[1]
		\BState {\textbf{Inputs}: $V_{c}$, $V_{\mathcal{N}}$}
		\BState {\textbf{Initialize}: }
		\For{$i \in V$} 
			\BState {\ \ \ \ $\mathbf{C}_i^0 =$ \emph{one-hot}(i);}
			\BState {\ \ \ \ $g_{s_i} = -\frac{1}{s_i}$;}
			\BState {\ \ \ \ $g_{s_{\mathcal{N}(i)}} = 0$;}
		\EndFor
		\BState {\textbf{Main}: }
		\For{$j\in V_c$}
			\For{$k < K_j$}
				\For{$i \in V_{\mathcal{N}_k}(j)$} 
				\BState	{\ \ \ \ \ \ \ \ \ \ \ \  $\mathbf{C}^t_i(j) = \sum_{\mathcal{N}(i)} s_{\mathcal{N}(i)} \cdot \mathbf{C}_{\mathcal{N}(i)}^{t-1}(j) + s_i\cdot \mathbf{C}_i^{t-1}(j)$;}
				\BState {\ \ \ \ \ \ \ \ \ \ \ \  $\mathcal{L}_c(i) = -\mathcal{M}_c(i)\cdot\ln (\mathbf{C}_i^t)$;}
				\BState {\ \ \ \ \ \ \ \ \ \ \ \  $g_{s_i} = - \frac{\mathcal{M}_c(i,j) \cdot \mathbf{C}_i^{t-1}(j)}{\hat{\mathbf{C}}_i^t(j)}$;}
				\BState {\ \ \ \ \ \ \ \ \ \ \ \  $g_{s_{\mathcal{N}(i)}} = - \frac{\mathcal{M}_c(i,j) \cdot \mathbf{C}_{\mathcal{N}(i)}^{t-1}(j)}{\mathbf{C}_i^{t}(j)} + g_{s_{\mathcal{N}(i)}}$;}
				\EndFor
			\EndFor
		\EndFor
	\BState {\textbf{Outputs}: $\mathbf{C}$ and $\frac {\partial \mathcal{L}_c}{\partial s_V}$.}
	\end{algorithmic}
\end{algorithm}

%% file: gps.tex
\begin{algorithm}
	\caption{Computing the object center of each node by Greedy Path Selection.}
	\label{alg:gps}
	\begin{algorithmic}[1]
		\BState {\textbf{Inputs}: $V$, $\mathbf{s}$.}
		\BState {\textbf{Initialize}: $\mathcal{F}$ = \emph{zeros}$(\frac{M}{D}, \frac{N}{D})$;}
		\For{$i \in V$} 
			\While{$\mathcal{F}(i) \neq i'$}
				\BState {\ \ \ \ \ \ \ \ $\mathcal{F}(i) = \argmaxA_{i'} \{s_i, s_{\mathcal{N}_i}\}$;}
			\EndWhile
		\EndFor
		\BState {Measure the distance between generated candidates $d_{ij}$;}
		\BState {For $d_{ij} < d_{T}$, run CP process on the minimal grid containing $v_i$ and $v_j$;}
		\BState {Update candidate list;}
		\BState {\textbf{Output}: $\mathcal{F}$.}
	\end{algorithmic}
\end{algorithm}

%% file: experiments.tex
\section{Experiments} \label{sect:exp}
\subsection{Datasets}

We evaluate the proposed model on three public scene text detection datasets, namely ICDAR 2013, ICDAR 2015 and MSRA-TD500, using the standard evaluation protocol proposed in \cite{wolf2006object,karatzas2015icdar,yao2012detecting}, respectively.

\textbf{SynthText \cite{gupta2016synthetic}} contains over 800,000 synthetic scene text images. They are created by blending natural images with text rendered with random fonts, sizes, orientations, and colors. It provides word level bounding box annotations. We only use this dataset to pre-train our model.

\textbf{ICDAR 2013 \cite{karatzas2013icdar}} is a dataset containing horizontal text lines. It has 229 text images for training and 223 images for testing.

\textbf{ICDAR 2015 \cite{karatzas2015icdar}} consists of 1000 training images and 500 testing images. This dataset features \emph{incidental} scene text images taken by Google Glasses without taking care of positioning, viewpoint and image quality.

\textbf{MARA-TD500 \cite{yao2012detecting}} is a multilingual dataset focusing on oriented texts. It consists of 300 training images and 200 testing images. 

\subsection{Experiment Details}

\begin{figure*}[]
	\centering
	\includegraphics[width=\linewidth]{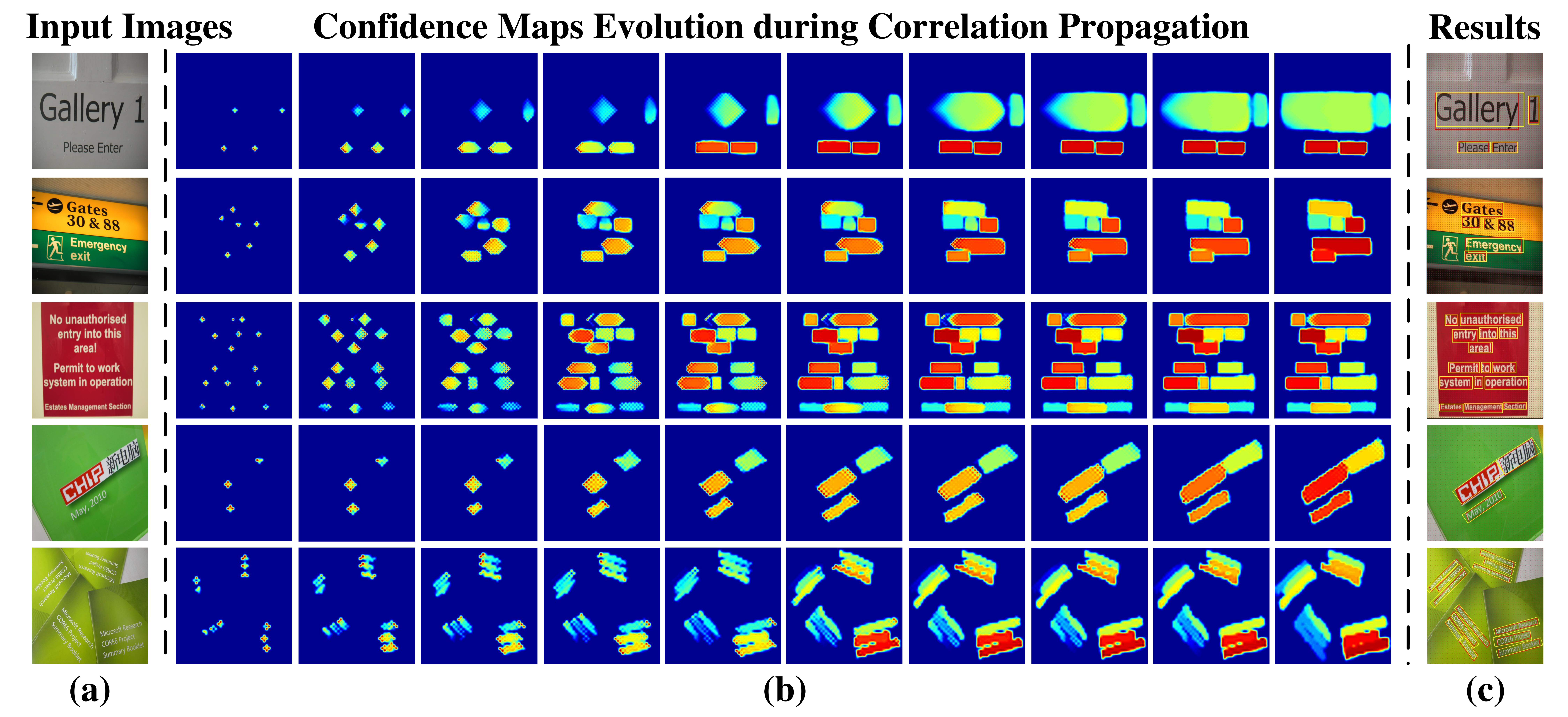}
	\caption{(a) Input images with different scales and orientations; (b) Confidence maps evolution in Correlation Propagation. From left to right, the object centers are firstly detected and according nodes belongs to the same object are progressively discovered; (c) Generated bounding box proposals. Our CPN framework can well model instance-level correlation of text objects, which provides addition information to detect long text objects by merging box candidates.}
	\label{fig:cp_demo}
\end{figure*}

The architecture of CPN is illustrated in Appendix, which consists of a backbone network, a Feature Pyramid Network (FPN) \cite{lin2016feature} and Correlation Propagation (CP) layer. The CNN backbone is optionally implemented by tailored (removing all fully-connected layers) VGG-16 \cite{simonyan2014very} and ResNet-50 \cite{he2016deep}. The deformable convolution is selectively inserted to the FPN.\par

The CPN model is pre-trained on SynthText dataset \cite{gupta2016synthetic} before fine-tuning on public real datasets. In pre-training phase, we apply AdaMax optimizer to train our model with a momentum of 0.9, learning rate decay factor of 0.99 per epoch, and the learning rate is set to $10^{-2}$ for first 10k iterations and turned to $10^{-3}$ for the rest 40k. In the fine-tuning phase, the learning rate initiated to $10^{-4}$ is reduced manually by a factor of 0.2 when saturation of loss is observed in the validation. For all dataset, we randomly crop the text regions and resize them to $512\times 512$ in the training phase.  The batch size is set to 20 and the workload is equally assigned to two GPUs. Data argumentation proposed in \cite{liu2016ssd} is applied to prevent overfitting. In the testing phase, the inputs are resized to the images with the same aspect ratio as the original images, but the short sides are fixed to 600 pixels. Both training and testing flows are implemented with Tensorflow \cite{abadi2016tensorflow} r1.1 on a workstation with Intel Xeon 8-core CPU (2.8GHz), 2 GTX 1080 Ti Graphics Cards, and 48GB memory. \par

\subsection{Detail Analysis}

\subsubsection{Visualizing Correlation Propagation}

In this experiment, we verify the behavior of Correlation Propagation by visualizing the evolution of confidence map $\mathbf{C}$. For the input images given in Fig. \ref{fig:cp_demo} (a), the corresponding $\mathbf{C}$ evolution is demonstrated (from left to right) in Fig. \ref{fig:cp_demo} (b). The heatmaps in (b) are produced by doing selective summation along the last dimension of $\mathbf{C}$. The 'selective' means that the summation only consists of specific slices $\mathbf{C}(j)$ which keep evolving during the CP process. Therefore, Fig. \ref{fig:cp_demo} (b) is showing the confidence evolution of all anchors on the potential object centers. As shown in Fig. \ref{fig:cp_demo} (b), the anchors located at the object centers are first activated. Subsequently, other anchors located within the text regions become activated inside-out during the CP process. Finally, the converged heatmaps (rightmost) reveal the correspondence between text objects and anchors and perfectly describe the geometry of each text instance.\par

As the spatial correlation of texts is well modeled by our CPN, one can adopt these information to assemble box proposals which significantly improve the performance of detecting long texts. Different from the existing R-CNN based scene text detection method \cite{yao2012detecting,jaderberg2016reading,huang2014robust,gupta2016synthetic,zhang2016multi,hu2017wordsup}, our method makes the most of the local information to provide a flexible detection flow without using pre-defined reference boxes. This feature could be very desirable when the statistics of object shape is not available.

\subsubsection{Ablation Experiment}

\input{cmp_baseline}
\input{perform_cmp_arch}

We conduct two ablation experiments to validate the effectiveness of CPN. One compares our method with a baseline method (Local-links) which also utilizes local correlation to predict instance-level bounding boxes. The other compares the performance between Faster R-CNN method \cite{ren2015faster} and CPN with different network implementations. All model are using VGG-16 as the backbone network and the number of parameters of the rest part is roughly equal. \par

In the first experiment, the Local-links predicts the foreground, background and four local links between nodes to capture the local correlation information. The instance-level bounding boxes are generated by finding the maximum connected component on the foreground regions. As shown in Table \ref{tab:baseline}, our method is overall better than the Local-links with 4$\%$ performance improvement. Since the detected segments are connected according to the intensity of local link scores, every node can only be aware of its local environment, which leads to unexpected merging between two individual objects. In contrast, due to the correlation propagation mechanism, all node have the global view of an image and can recognize its center by collecting local correlation from other nodes, which greatly improves the robustness as well as flexibility. \par

In the second experiment, CPN method overall outperforms the Faster R-CNN methods, as shown in Tab. \ref{tab:cmp:arch}. It is observed that Faster R-CNN fails when detecting text object with an extremely large aspect ratio. Although the deformable convolution can improve the performance by around 2$\%$, it is still not comparable with CPN method that utilizes local correlation. Our CPN works extremely well in detecting oriented texts. On MSRA-TD500, our method significantly outperforms the Faster R-CNN method by up to 10$\%$ since the geometry information captured by CPN is introduced to improve orientation prediction. On ICDAR 2013 and ICDAR 2015, our models also achieve better performance than the baseline models.\par

\subsubsection{Impact of Down-sampling Factor $D$}
We evaluate the performance and GPU running time of our method under different settings of $D$ on IC13 and IC15 shown in Table \ref{tab:exp_ic}. On IC13, the performance is not greatly affected by the prediction density since IC13 is mainly composed of large and middle-sized texts. However, increasing the prediction density can boost the performance on IC15, as it is dominated by middle and small texts. Moreover, the running time of CP and GPS are insensitive to $D$ due to the high parallelism of GPUs. The total running time is also insensitive to $D$, as $D$ is increased by stacking additional up-sampling layer. In conclusion, the CPN achieve much better performance with higher prediction density but a slight drop in speed.

\input{den_acc}

\subsubsection{Direct Correlation Propagation v.s. Greedy Path Selection} \label{sect:cp_vs_gps}

\input{cmp_infer}

\input{cmp_exist_works}

Table \ref{tab:cmp_infer} compares the performance between our direct CP process and our improved GPS algorithm on three public datasets. On ICDAR 2013 and MSRA-TD500 dataset, GPS has roughly the same performance as the CP algorithm with the precision of 87.9, recall of 87.5 and F-score of 87.7. On ICDAR 2015, GPS algorithm performs slightly worse than CP algorithm with the precision of 69.6, recall of 74.5 and F-score of 74.5.  In terms of inference time, the GPS algorithm runs much faster ($24\times$) than the CP algorithm due to its high level of parallelism. It concludes that the GPS algorithm is more efficient while maintaining comparable performance.

\subsubsection{Performance Evaluation}

\begin{figure}[]
	\centering
	\includegraphics[width=\linewidth]{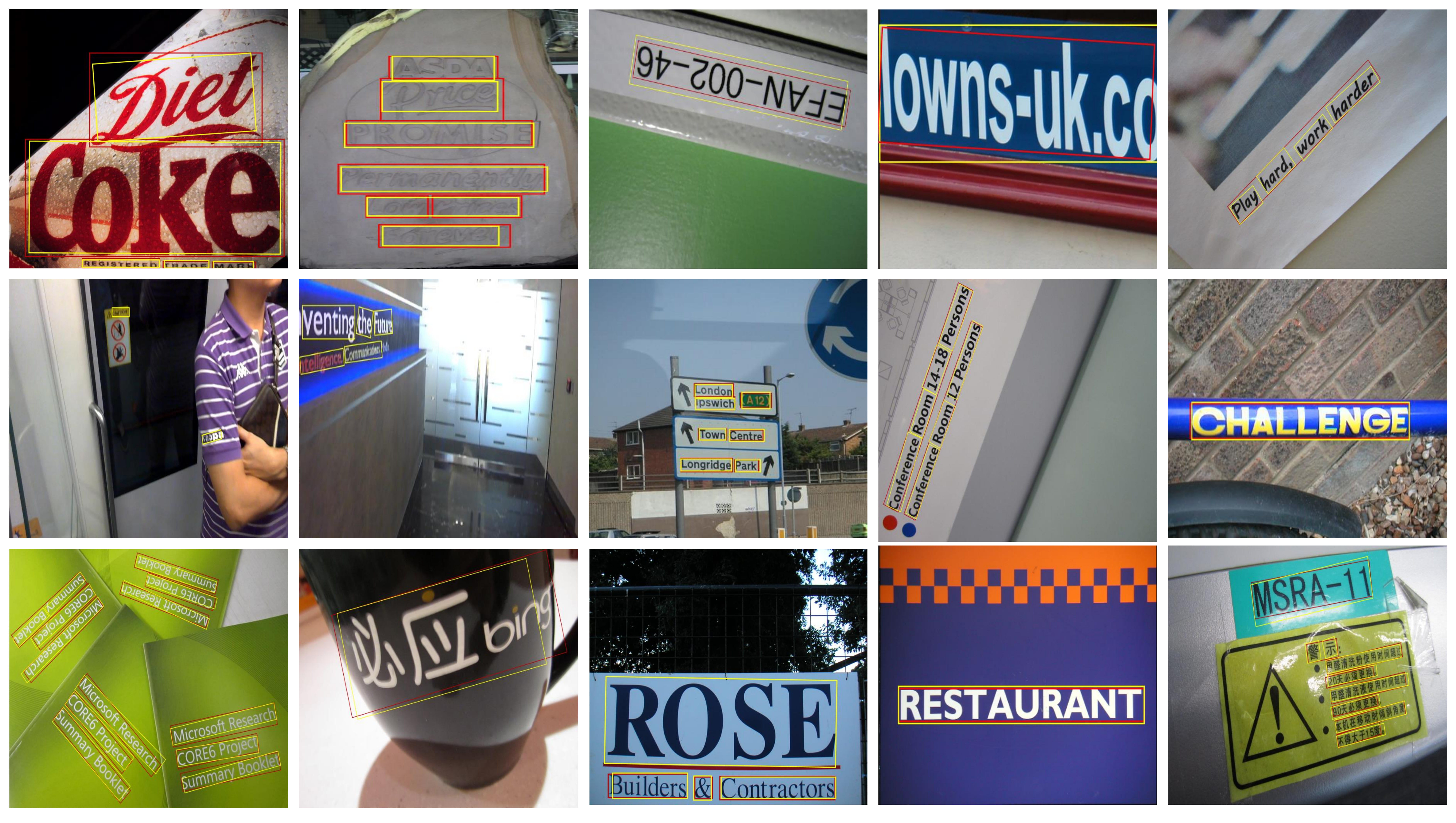}
	\caption{Example resutls on ICDAR 2013, ICDAR 2015 and MSRA-TD 500 datasets. The predicted bounding boxes by CPN are labeled in yellow and the ground-truth bounding boxes are marked in red. Our CPN method can effectively detect texts with various aspect ratios and orientations.}
	\label{fig:loc_results}
\end{figure}

Table \ref{tab:full_cmp} compares CPN with the published works on public datasets. On ICDAR 2013, CPN has competitive performance (precision: 88.1, recall: 90.5 and F-score: 89.3) when compared with the state-of-art work \cite{hu2017wordsup}, but enjoys much higher inference speed (0.50s v.s. 0.09s). On ICDAR 2015, our method achieves a new state-of-art recall rate (82.6) but perform slightly worse than work \cite{he2017deep} in F-score due to lower precision. It is notable that CPN outperforms the existing methods by a great margin on MSRA-TD500 (precision: 89.8, recall: 82.7 and F-score: 86.1). MSRA-TD500 is a challenging dataset consisting of large amounts of muli-oriented text objects with large aspect ratios. Our method performs extremely well in handling oriented text objects. The bounding box prediction results are demonstrated in Fig. \ref{fig:loc_results}.

\vspace{0pt}

%% file: cmp_baseline.tex
\begin{table*}[t]\renewcommand{\arraystretch}{1}
	\centering
	\caption{Comparison between CPN and baseline model (Local-links) on public benchmarks. It shows that the our CPN method performs better than the Local-links method.}
	\label{tab:baseline}
	\begin{tabular*}{\textwidth}{@{\extracolsep{\fill}}|l|c|c|c|c|c|c|c|c|c|}
		\hline
		\textbf{Datasets} & \multicolumn{3}{c|}{\textbf{ICDAR 2013}} & \multicolumn{3}{c|}{\textbf{ICDAR 2015}} & \multicolumn{3}{c|}{\textbf{MSRA-TD500}} \\ \hline
		\textbf{Methods}  & \textbf{P}   & \textbf{R}  & \textbf{F}  & \textbf{P}   & \textbf{R}  & \textbf{F}  & \textbf{P}   & \textbf{R}  & \textbf{F}  \\ \hline
		Local-links (VGG-16)       & 82.3         & 85.1        & 83.4        & 61.6             & 76.9            &  68.4           & 83.3         & 70.3        & 76.2   \\
		CPN (VGG-16)              & \textbf{85.6}         & \textbf{86.9}        & \textbf{86.2 }       & \textbf{69.8}         & \textbf{80.7}        & \textbf{74.8}            & \textbf{87.6}         & \textbf{75.3}        & \textbf{80.9}        \\ \hline
	\end{tabular*}
\end{table*}

%% file: perform_cmp_arch.tex
\begin{table*}[t]\renewcommand{\arraystretch}{1}
	\centering
	\caption{Performance comparison between CPN and general object detection method in detecting scene texts. Our CPN model perform better especially in detecting long and oriented texts.}
	\label{tab:cmp:arch}
	\begin{tabular*}{\textwidth}{@{\extracolsep{\fill}}|l|c|c|c|c|c|c|c|c|c|}
		\hline
		\textbf{Datasets}        & \multicolumn{3}{c|}{\textbf{ICDAR 2013}} & \multicolumn{3}{c|}{\textbf{ICDAR 2015}} & \multicolumn{3}{c|}{\textbf{MSRA-TD500}} \\ \hline
		\textbf{Architectures}   & \textbf{P}   & \textbf{R}  & \textbf{F}  & \textbf{P}   & \textbf{R}  & \textbf{F}  & \textbf{P}   & \textbf{R}  & \textbf{F}  \\ \hline
		Faster R-CNN (ResNet-50)             & 80.7        & 86.2       & 83.4        & 76.4      & 73.9        & 75.1             & 72.9         & 72.7        & 72.8   \\
		Faster R-CNN (ResNet-50+Deform)  & 81.4         & 86.8        & 84.0        & 77.6         & 75.3        & 76.4             & 75.7         & 77.4        & 76.5   \\
		CPN(VGG-16+FPN)               & 85.6         & 86.9        & 86.2        & 69.8         & 80.7        & 74.8             & 87.6         & 75.3        & 80.9   \\ 
		CPN(ResNet-50+FPN)            & 83.2         & 90.1        & 86.6        & 75.7         & 82.0        & 78.7             & 87.1         & 80.4        & 83.6   \\ 
		CPN(ResNet-50+Deform FPN)     & \textbf{88.1}         & \textbf{90.5}        & \textbf{89.3}        & \textbf{76.9}         & \textbf{82.6}        & \textbf{79.6}        & \textbf{89.8}         & \textbf{82.7}        & \textbf{86.1}   \\ \hline
	\end{tabular*}
\end{table*}

%% file: den_acc.tex
\begin{table}[h]\renewcommand{\arraystretch}{1.0}
	\centering
	\small
	\caption{Analysis experiment on ICDAR 2013 and ICDAR 2015. Increasing down-sampling factor can lead to performance improvement on ICDAR 2015 since it is dominated by small texts. The GPS accelerated inference algorithm is much faster than the pure CP inference algorithm.}
	\label{tab:exp_ic}
	\begin{tabular}{|l||c|c|c|c|c|c|}
		\hline
		& \textbf{P}    & \textbf{R}    & \textbf{F} & \multicolumn{1}{c|}{\begin{tabular}[c]{@{}c@{}}CP\\ (ms)\end{tabular}} & \multicolumn{1}{c|}{\begin{tabular}[c]{@{}c@{}}GPS\\ (ms)\end{tabular}} & \multicolumn{1}{c|}{\begin{tabular}[c]{@{}c@{}}test\\ (ms)\end{tabular}} \\ \hline
		\multicolumn{1}{|c||}{\begin{tabular}[c]{@{}c@{}}IC13\\ D=16\end{tabular}}  & 88.15 & 90.52 & 89.30 & 0.24 & 0.012 & 0.09\\ \hline
		\multicolumn{1}{|c||}{\begin{tabular}[c]{@{}c@{}}IC13\\ D=8\end{tabular}}   & 87.90 & 90.36 & 89.11 & 0.25 & 0.015 & 0.11 \\ \hline
		\multicolumn{1}{|c||}{\begin{tabular}[c]{@{}c@{}}IC13\\ D=4\end{tabular}}   & 88.60 & 89.91 & 89.25 & 0.29 & 0.021 & 0.16 \\ \hline
		\multicolumn{1}{|c||}{\begin{tabular}[c]{@{}c@{}}IC15\\ D=16\end{tabular}}   & 76.90 & 82.61 & 79.65 & 0.23 & 0.011 & 0.09 \\ \hline
		\multicolumn{1}{|c||}{\begin{tabular}[c]{@{}c@{}}IC15\\ D=8\end{tabular}}  & 83.90 & 79.62 & 81.70 & 0.24 & 0.018 & 0.12\\ \hline
		\multicolumn{1}{|c||}{\begin{tabular}[c]{@{}c@{}}IC15\\ D=4\end{tabular}}  & 89.25 & 80.12 & 84.44 & 0.27 & 0.022 & 0.15 \\ \hline

	\end{tabular}
\end{table}

%% file: cmp_infer.tex
\begin{table*}[] \renewcommand{\arraystretch}{1}
	\centering
	\caption{Direct Correlation Propagation v.s. Greedy Path Selection. The GPS accelerated inference runs much faster than pure CP approach while does not affect the accuracy.}
	\label{tab:cmp_infer}
	\begin{tabular*}{\textwidth}{@{\extracolsep{\fill}}|l|c|c|c|c|c|c|c|c|c|c|}
		\hline
		\textbf{Datasets}       & \multicolumn{4}{c|}{\textbf{ICDAR 2013}} & \multicolumn{3}{c|}{\textbf{ICDAR 2015}} & \multicolumn{3}{c|}{\textbf{MSRA-TD500}} \\ \hline
		\textbf{Infer Approach} & \textbf{P}   & \textbf{R}  & \textbf{F} & \textbf{time (ms)} & \textbf{P}   & \textbf{R}  & \textbf{F}  & \textbf{P}   & \textbf{R}  & \textbf{F}  \\ \hline
		Correlation Propagation       & \textbf{88.4}         & 87.1        & \textbf{87.7}   &  0.24   & \textbf{69.8}         & \textbf{80.7}        & \textbf{74.8}        & \textbf{87.6}         & 75.3        & \textbf{80.9}        \\ 
		Greedy Path Selection      & 87.9         & \textbf{87.5}        & \textbf{87.7}    &  \textbf{0.01}  & 69.6         & 80.1        & 74.5             & 87.7         & \textbf{75.9}        & \textbf{80.9}            \\ \hline
	\end{tabular*}
\end{table*}

%% file: cmp_exist_works.tex
\begin{table*}[t]\renewcommand{\arraystretch}{1}
\centering
\caption{Localization performance on ICDAR-13, ICDAR-15 and MSRA-TD500.}
\label{tab:full_cmp}
\scalebox{0.99}{
\begin{tabular*}{\textwidth}{@{\extracolsep{\fill}}|l||ccc|c||ccc||ccc|}
	\hline
	Dataset                                                 & \multicolumn{4}{c||}{ICDAR-13}                 & \multicolumn{3}{c||}{ICDAR-15}                 & \multicolumn{3}{c|}{MSRA-TD500}               \\ \hline
	Methods                           & \multicolumn{1}{c}{P}             & \multicolumn{1}{c}{R}             & \multicolumn{1}{c|}{F}    & \multicolumn{1}{c||}{time(s)}        & \multicolumn{1}{c}{P}             &\multicolumn{1}{c}{R}             & \multicolumn{1}{c||}{F}             & \multicolumn{1}{c}{P}             & \multicolumn{1}{c}{R}             & \multicolumn{1}{c|}{F}             \\ \hline
	TextFlow \cite{tian2015text}                          & 85.1          & 76.4          & 80.6    & 0.94     & -             & -             & -             & -             & -             & -             \\ 
	Jaderberg \emph{et al.} \cite{jaderberg2016reading} & 89.2          & 68.3          & 77.4      & 1.00   & -             & -             & -             & -             & -             & -             \\ 
	Zhang \emph{et al.} \cite{zhang2016multi}             & 88.4          & 78.4          & 83.2    & 60.0     & 71.1          & 43.1          & 53.7          & 83.2          & 67.4          & 74.5          \\ 
	Gupta \emph{et al.} \cite{gupta2016synthetic}       & 92.0          & 75.3          & 82.8   & 0.06      & -             & -             & -             & -             & -             & -             \\ 
	Yao \emph{et al.} \cite{yao2016scene}               & -             & -             & -      & -      & 72.4          & 59.0          & 65.0          & 77.2          & 75.0          & 76.1          \\ 
	TextBox \cite{liao2017textboxes}                      & 88.4          & 83.0          & 85.6  & 0.05       & -             & -             & -             & -             & -             & -             \\ 
	CTPN \cite{tian2016detecting}                         & 92.7 & 82.9          & 87.5 & 1.40 & 52.3       & 74.6          & 62.3          & -             & -             & -             \\ 
	SegLink \cite{shi2017detecting}                       & 87.7          & 83.0          & 85.3   & 0.05      & 73.1 & 76.8 & 75.0 & 86.0          & 70.0          & 77.0          \\ 
	DeepReg \cite{he2017deep}      & 92.0              &81.0           &86.0   & 0.90    & 82.0   & 80.0     & 81.0      & 77.0     & 70.0    & 74.0 \\ 
	WordSup \cite{hu2017wordsup}      & \textbf{93.3}              & 87.5           & \textbf{90.3}  &  0.50    & 79.3    & 77.0     & 78.1      & -     & -    & - \\ 
	He \emph{et al.} \cite{gomez2018single} & 88.0 & 87.0 & 88.0 & - & 84.0 & 83.0 & 83.0 & - & - & -\\ 
	Lyu \emph{et al.} \cite{lyu2018multi} & 92.0 & 84.4 & 88.0 & 1.00 & 89.5 & 79.7 & 84.3 & - & - & -\\ \hline
	CPN (D=16)     & 88.1         & \textbf{90.5}        & 89.3    & 0.09   & 76.9         & \textbf{82.6}        & 79.6        & \textbf{89.8}         & \textbf{82.7}        & \textbf{86.1}   \\
	CPN (D=8)  & 87.9 & 90.4 & 89.1 & 0.12 & 83.9 & 79.6 & 81.7 & - & - & - \\ 
	CPN (D=4)  & 88.6 & 89.9 & 89.3  &0.16 & \textbf{89.3} & 80.1 & \textbf{84.4} & - & - & -\\\hline

\end{tabular*}
}
\end{table*}

%% file: conclusion.tex
\section{Conclusion}

In this work, we have developed a Correlation Propagation mechanism which can represent object instance in a distributed manner. We combine this mechanism into advanced deep neural networks and build an end-to-end trainable model for scene text detection. It can learn the local correlations and uses them to construct instance-level representations by node-to-node communication. Our method is flexible and robust to text with various aspect ratios and orientations. It achieves the state-of-art performance on public benchmarks.\par

%% file: acks.tex
\section{Acknowlegements}
We gratefully acknowledge the support of NVIDIA Corporation with the donation of the GPU used for this research.

%% file: appendix.tex
\clearpage
\appendix
\section{Appendices}

\begin{figure}[h]
	\centering
	\includegraphics[width=0.9\linewidth]{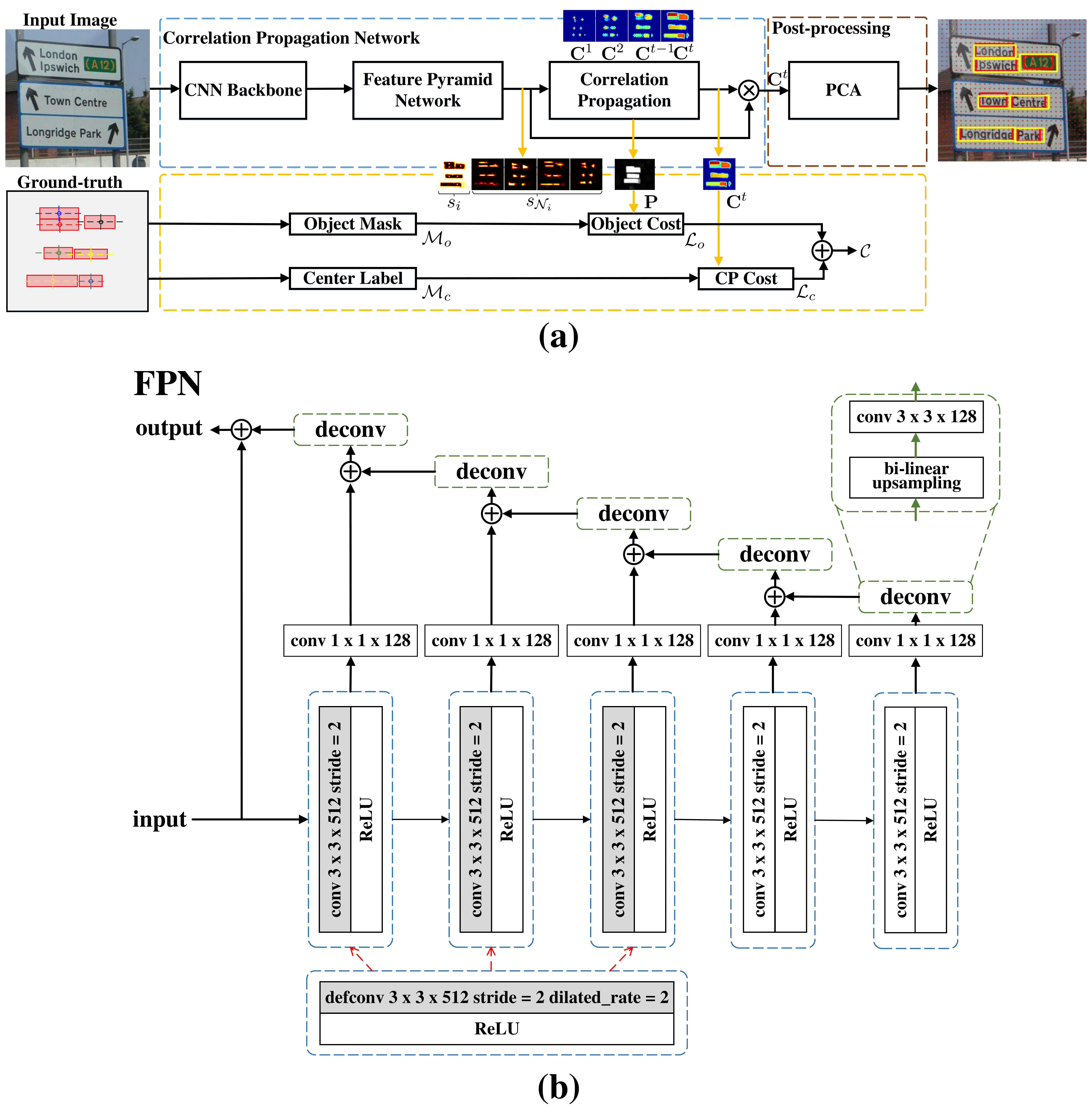}
	\caption{(a) Work flow of training and inference of Correlation Propagation Networks; Architecture of Feature Pyramid Network (FPN) used in our method with optional deformable convolution.}
	\label{fig:sys_flow}
\end{figure}

The architecture of CPN, inference flow and training flow are illustrated in Fig. \ref{fig:sys_flow} (a). A CPN consists of a CNN backbone network. All the fully-connected layers are removed, resulting in an output resolution of $1/16$. Here, we apply VGG-16 \cite{simonyan2014very} and ResNet-50 \cite{he2016deep} alternatively. The output of the backbone network is subsequently fed to a Feature Pyramid Network (FPN) \cite{lin2016feature}. As shown in Fig. \ref{fig:sys_flow} (b), deformable convolution is optionally applied to the downsample layer of FPN which enables an adaptive data sampling on an image. The FPN fuses multiscaled features into a fixed-size feature map where we make a prediction of $s_i$, $s_{\mathcal{N}_i}$ and $\mathbf{P}_i$ for each node with $1\times 1$ convolution window. Note that the FPN is flexible to predict $\mathbf{P}$ and $\mathbf{C}$ is the different resolution by manipulating the number of upsampling layers. Therefore, our architecture can make a prediction in an output resolution even higher than $1/16$ ($1/8$, $1/4$. etc.).

